\documentclass[11pt]{article}
\usepackage[preprint]{acl}
\usepackage{times}
\usepackage{latexsym}
\usepackage[T1]{fontenc}
\usepackage[utf8]{inputenc}
\usepackage{microtype}
\usepackage{subcaption}
\usepackage{inconsolata}
\usepackage{graphicx}
\usepackage{xcolor}
\usepackage{amssymb}
\usepackage{amsmath}
\usepackage{booktabs}
\usepackage{multirow}
\usepackage{listings}

\title{From Bounding Boxes to Visual Reasoning: \\ An On-Policy Data Annotation Tool for Vision-Language Models}

\author{
\textbf{Like Zhang\textsuperscript{1}\thanks{\texttt{zhanglk25@mails.jlu.edu.cn}}},
\textbf{Runliang Niu\textsuperscript{1}},
\textbf{Shiqi Wang\textsuperscript{1}},
\textbf{Xiyu Hu\textsuperscript{1}},
\textbf{Qianli Xing\textsuperscript{2}},
\\
\textbf{Pan Wang\textsuperscript{3}},
\textbf{Qingzu He\textsuperscript{3}},
\textbf{Qi Wang\textsuperscript{1}\thanks{Corresponding author. \texttt{qiwang@jlu.edu.cn}}}
\\
\textsuperscript{1}School of Artificial Intelligence, Jilin University
\\
\textsuperscript{2}College of Computer Science, Jilin University 
\textsuperscript{3}OPPO
}

\begin{document}

\maketitle
\begin{abstract}
Vision-language models (VLMs) are rapidly advancing toward sophisticated grounded structured visual reasoning. Training models for such advanced capabilities demands a new genre of data that seamlessly unifies spatial coordinates, open-vocabulary descriptions, structured attributes, and topological relationships into a singular representation. However, existing data annotation tools fundamentally fail to meet these intricate demands, suffering from three systematic bottlenecks: limited expressiveness, severe annotation-training decoupling, and poor data reusability. 
To bridge this infrastructure gap, we introduce an open-source annotation tool,  \textsc{ScreenAnnotator}. First, we define a unified annotation atom schema that binds spatial, semantic, and structural primitives into a single unit. Second, we implement an on-policy annotation loop embedded with a Bayesian Annotation Verifier (BAV). Finally, we design a template-driven multi-task data synthesis process dynamically transforms static atoms into diverse multi-dimensional reasoning tasks, eliminating redundant re-annotation. 
The on-policy loop drives the annotation accept rate to nearly 100\% on flowcharts and 77\% on GUI screenshots, while steadily reducing per-image annotation time as labeled data accumulate. In the flowchart scenario, fine-tuning a VLM yields 76.1\% average accuracy, which is a 35.1\% point absolute gain.
Our code is available at: \url{https://github.com/WnQinm/Annotator}.

\end{abstract}

\section{Introduction}

Vision-language models (VLMs) have shifted from basic visual perception to grounded structured visual reasoning. In complex tasks such as GUI understanding and flowchart parsing, modern VLMs perform multi-step reasoning by weaving spatial markers into their textual chains of thought. Training these models therefore requires data that unifies spatial coordinates, open-vocabulary descriptions, structured attributes, and topological relationships into a single representation.

\begin{figure}[t]
    \centering
    \includegraphics[width=\linewidth]{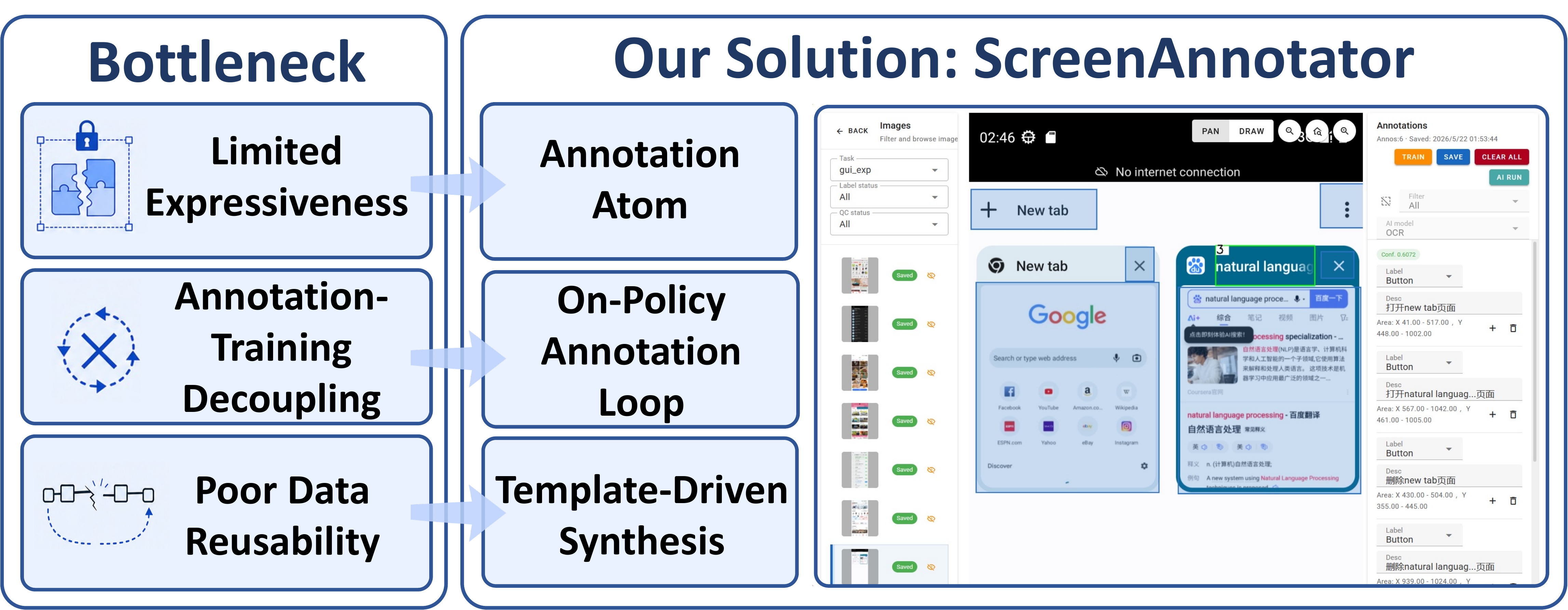}
    \caption{Motivation of \textsc{ScreenAnnotator}: bridging the infrastructure gap between conventional image annotation tools and the data demands of modern VLMs.}
    \label{fig:motivation}
    \vspace{-1em}
\end{figure}

Recent years have witnessed notable advancements in image annotation platforms, evolving from manual desktop applications to collaborative, web-based systems integrated with vision foundation models~\citep{Sager03072021}. We provide a systematic comparison of these representative open-source and commercial tools in Table~\ref{tab:tool-comparison}. However, these contemporary tools fundamentally fail to meet the intricate data demands of modern VLMs as they remain anchored in the closed-set detection paradigm, as shown in Figure~\ref{fig:motivation}. Existing platforms suffer from three systematic bottlenecks: (1) limited expressiveness, failing to jointly bind spatial locations, free-form text, and attributes; (2) annotation-training decoupling, relying on linear offline pipelines without interactive, on-policy feedback or self-supervised quality control; and (3) poor data reusability, requiring redundant re-annotation for diverse downstream tasks. This lagging infrastructure has become a critical barrier hindering the scalable curation of high-quality data for advanced VLM reasoning.

To bridge this infrastructure gap, we introduce \textsc{ScreenAnnotator}, an open-source annotation tool specifically engineered for VLM training data. To address limited expressiveness, we define a unified annotation atom schema that jointly binds spatial location, categorical identity, free-form description, and structured attributes within a single unit. 

To eliminate annotation-training decoupling, we implement an on-policy annotation loop. Our framework iteratively alternates between model-assisted pre-annotation, human correction, and immediate model retraining. This closed-loop co-evolution progressively reduces the human correction burden and yields quantitative performance emergence curves for early stopping. To ensure data integrity, we embed a Bayesian Annotation Verifier (BAV) that estimates per-annotation posterior uncertainty and routes high-risk instances for human re-review, providing self-supervised quality control without manual quality labels. 

To improve data reusability, we propose a template-driven multi-task data synthesis procedure. A single set of annotation atoms can be dynamically converted into diverse VLM reasoning tasks such as attribute lookup, relational reasoning, and spatial inference via schema templates. This approach decouples data diversity from annotation cost, drastically increasing the marginal supervision of each annotation without new instance-level labeling.

In summary, our primary contributions to the community are threefold:

\noindent\textbullet\ \textbf{ScreenAnnotator.} We present a novel open-source annotation tool that extends the conventional closed-set detection paradigm. We define a unified annotation atom schema that carries rich semantic information and supports instance-level referencing.

\noindent\textbullet\ \textbf{On-Policy Annotation Loop.} During annotation, assistive models are rapidly updated on-policy, while a Bayesian Annotation Verifier (BAV) detects potential errors. In human-in-the-loop experiments, we observe a clear positive feedback in annotation efficiency, with both the amount of human correction and correction time decreasing after each model update.

\noindent\textbullet\ \textbf{Reusable Annotation Effort.} We propose a template-driven multi-task data synthesis procedure that converts a one-time annotation effort into reusable multi-task VLM training data. In the flowchart QA scenario, the fine-tuned VLM achieves 76.1\% average accuracy, an absolute gain of 35.1 percentage points over the base model, with consistent improvements across all task types.

\section{Related Work}

\begin{table*}[t]
\centering
\fontsize{8pt}{8pt}\selectfont
\setlength{\tabcolsep}{2.5pt}
\renewcommand{\arraystretch}{1.1}
\caption{
  Comparison of representative annotation tools along five dimensions relevant to VLM training data.
  Most existing tools target closed-set detection formats (COCO, VOC, YOLO), perform model updates outside the annotation tool, and rely on manual review or inter-annotator agreement for quality control.
  None natively produces the grounded, open-vocabulary formats required by modern VLMs.
  \checkmark~= natively supported;\;
  $\triangle$~= partial or enterprise-only;\;
  $\times$~= not supported.
  \textsuperscript{†}~Enterprise edition required.
}
\label{tab:tool-comparison}
\begin{tabular}{p{1.9cm} p{1.7cm} p{3.0cm} p{2.7cm} p{2.6cm} p{3.0cm}}
\toprule
\textbf{Tool} & \textbf{License} & \textbf{Model-Assisted Labeling} & \textbf{Online / Incr.\ Training} & \textbf{Quality Control} & \textbf{Export Data for VLM} \\
\midrule
LabelImg 
  & Open-source
  & $\times$
  & $\times$
  & $\times$
  & $\times$ \newline (VOC\,XML / YOLO) \\

LabelMe \cite{russell2008labelme}
  & Open-source
  & $\triangle$ \newline (SAM click-to-segment; commercial only)
  & $\times$
  & $\times$
  & $\times$ \newline (per-image JSON) \\

CVAT \cite{sekachev2019computer}
  & Open-source\textsuperscript{†}
  & \checkmark \newline (Nuclio; custom auto-annotation)
  & $\times$ \newline (static weights; no retraining)
  & $\triangle$\textsuperscript{†} \newline (review workflow, annotator analytics)
  & $\times$ \newline (COCO / VOC / YOLO only) \\

Label Studio \cite{labelstudio}
  & Open-source\textsuperscript{†}
  & \checkmark \newline (ML Backend; GroundingDINO, SAM)
  & $\triangle$\textsuperscript{†} \newline (submission-triggered; not on-policy)
  & $\triangle$\textsuperscript{†} \newline (IAA, review workflow)
  & $\times$ \newline (generic formats; no grounding or captioning) \\

Roboflow
  & Commercial
  & \checkmark \newline (Autodistill: GroundingDINO + SAM2)
  & $\triangle$ \newline (manual trigger for next-round Label Assist)
  & $\triangle$ \newline (dataset analytics, annotation history)
  & $\times$ \newline (COCO / YOLO; no bbox+text grounding) \\

V7 Darwin
  & Commercial
  & \checkmark \newline (SAM-based pixel-perfect auto-annotation)
  & $\times$ \newline (no in-session update)
  & \checkmark \newline (multi-stage review, consensus scoring)
  & $\times$ \newline (COCO / VOC; no VLM captioning) \\

Supervisely
  & Commercial
  & \checkmark \newline (one-click NN labeling with deployed models)
  & $\triangle$ \newline (active learning; manual orchestration)
  & \checkmark \newline (QA stats, heatmaps, annotator performance)
  & $\times$ \newline (COCO; no grounding format)\\ 
Prodigy
  & Commercial
  & \checkmark \newline (model-in-the-loop, binary accept / reject)
  & $\triangle$ \newline (via \texttt{spacy train}; not real-time)
  & $\times$ \newline (single-annotator; no multi-annotator QC)
  & $\times$ \newline (spaCy / JSONL; no VLM task format) \\

\midrule
\textbf{Ours}
  & Open-source
  & \checkmark \newline (co-evolving on-policy model inference)
  & \checkmark \newline (automatic continuous on-policy retraining)
  & \checkmark \newline (Bayesian, bbox + token)
  & \checkmark \newline (bbox + description + attributes; grounding / VQA) \\
\bottomrule
\end{tabular}
\end{table*}

\subsection{Image Annotation Tools and Platforms}

Table~\ref{tab:tool-comparison} surveys representative annotation tools\footnote{
LabelImg \url{https://github.com/tzutalin/labelImg}\\
Roboflow \url{https://github.com/roboflow}\\
V7 Darwin \url{https://github.com/v7labs}\\
Supervisely \url{https://github.com/supervisely}\\
Prodigy \url{https://prodi.gy/}} along four dimensions: whether they support model-assisted pre-annotation, online model retraining from accumulated annotations, automatic annotation error detection for quality control, and export formats directly consumable for VLM training. Most existing platforms are primarily designed for conventional object detection workflows, where annotations mainly consist of categorical labels and bounding boxes for detector models such as YOLO ~\citep{yolo11_ultralytics}. Although some platforms have recently integrated vision foundation models for pre-labeling, they still lack automatic on-policy retraining, uncertainty-driven quality control, and native VLM data export.

\subsection{On-Policy Annotation and Quality Estimation}

Deep active learning selects informative samples via uncertainty, diversity, or hybrid acquisition strategies~\cite{zhan2022comparative,ren2022survey}, with task-specific extensions to object detection~\cite{haussmann2020scalable,elezi2022not}. SAM's data engine~\cite{kirillov2023segment} is the closest precedent to an on-policy loop, alternating model-generated masks, human correction, and retraining. On the quality side, crowdsourcing aggregation methods~\cite{zheng2017truth} estimate annotator reliability from multi-annotator redundancy. Noisy-label methods such as DivideMix~\cite{li2020dividemix} and SOP+~\cite{liu2022robust_sop} tolerate label errors at training time but do not prevent them during annotation. 
Bayesian uncertainty estimation~\cite{lakshminarayanan2017simple,harakeh2020bayesod} offers a principled quality signal but has so far targeted prediction rather than annotation verification.

\subsection{VLM Training Data and Grounded Visual Reasoning}
\label{sec:vlm-data}

The data demands of VLMs have evolved through three stages. 
The first generation aligned models with web-crawled image caption pairs~\citep{sharma2018conceptual,changpinyo2021cc12m,schuhmann2022laion}. This regime enabled global captioning and visual question answering capabilities, but left models weak at localization. GPT-4V, for instance, fails on a substantial fraction of GUI grounding queries~\citep{niu2024screenagent}. A second stage, driven by visual agents that demand richer reasoning, trained VLMs on object detection datasets to emit coordinates directly as text tokens, enabling detection and grounding~\citep{chen2021pix2seq,chen2022unified,chen2023shikra,peng2023kosmos2}. The latest trend moves spatial primitives from terminal outputs into intermediate reasoning steps. Points and bounding boxes serve as visual anchors that reduce hallucination and improve verifiability~\citep{mitra2023ccot,shao2024visualcot,fan2025grit,deepseek2026tvp}. These trends indicate that VLMs require increasingly diverse annotation structures, yet existing annotation tooling has not kept pace.
\section{\textsc{ScreenAnnotator}}

\begin{figure*}
    \centering
    \includegraphics[width=\textwidth]{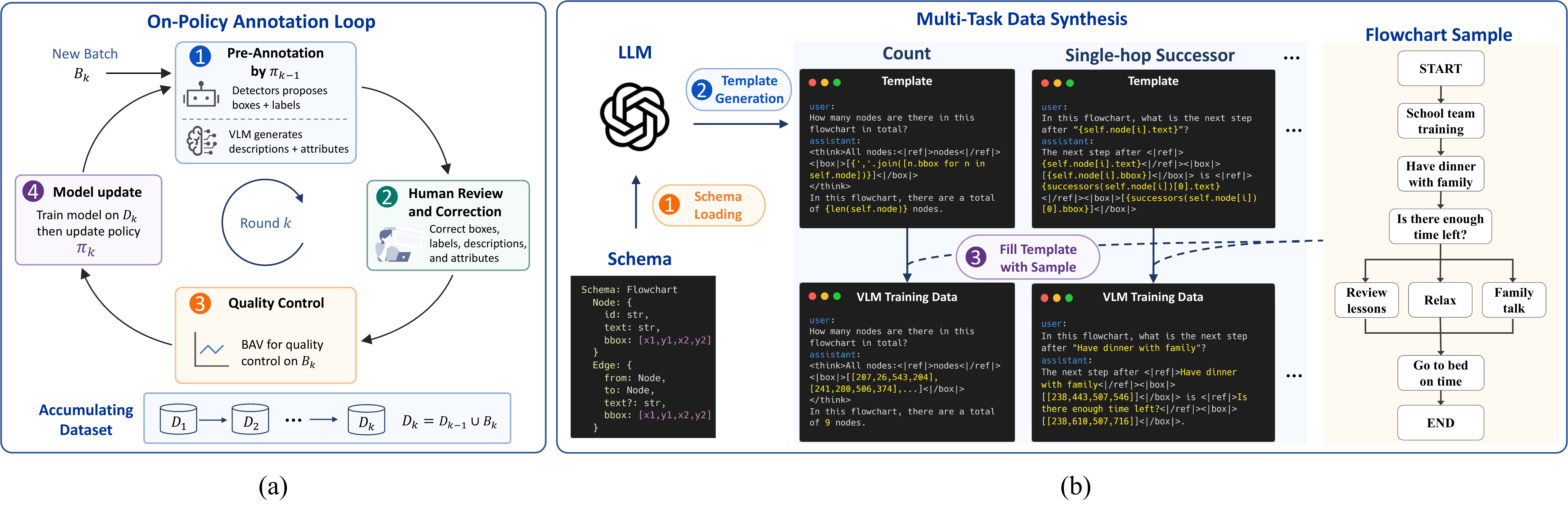}
    \caption{(a) The On-Policy Annotation Loop: At each round, (1) the current policy model (a detector + VLM) pre-annotates a new image batch; (2) a human annotator reviews and corrects the predictions; (3) a Bayesian verifier flags suspicious annotations for a second review pass; (4) both models are retrained on the accumulated quality-controlled dataset and deployed in the next round. (b) Multi-Task Data Synthesis. We take flowchart parsing as an example. A domain-specific schema declaring the node and edge structure is first loaded into an LLM. The LLM then expands the schema into a diverse set of conversation templates. Each template carries placeholders bound to schema fields, together with grounding markers (<ref>, <box>) that tie textual answers to spatial evidence. Every template is then instantiated against every annotated sample, so a single labeled image is reused across all compatible templates to produce multi-task VLM training data from one labeling effort.}\label{fig:framework}
    \vspace{-1em}
\end{figure*}

First, we define the Unified Annotation Atom, which binds spatial coordinates, open-vocabulary descriptions, and structured attributes into a single representation.
Second, we design an On-Policy Annotation Loop to construct the atom dataset. The loop embeds a Bayesian Annotation Verifier (BAV) for self-supervised quality control.
Finally, we design a Template-Driven Data Synthesis module that converts annotation atoms into diverse VLM reasoning tasks, including attribute lookup, path traversal, and spatial inference, enabling a single labeling effort to support multiple downstream tasks without re-annotation.

\subsection{Unified Annotation Atom Schema}

\begin{figure}[t]
    \centering
    \includegraphics[width=0.48\textwidth]{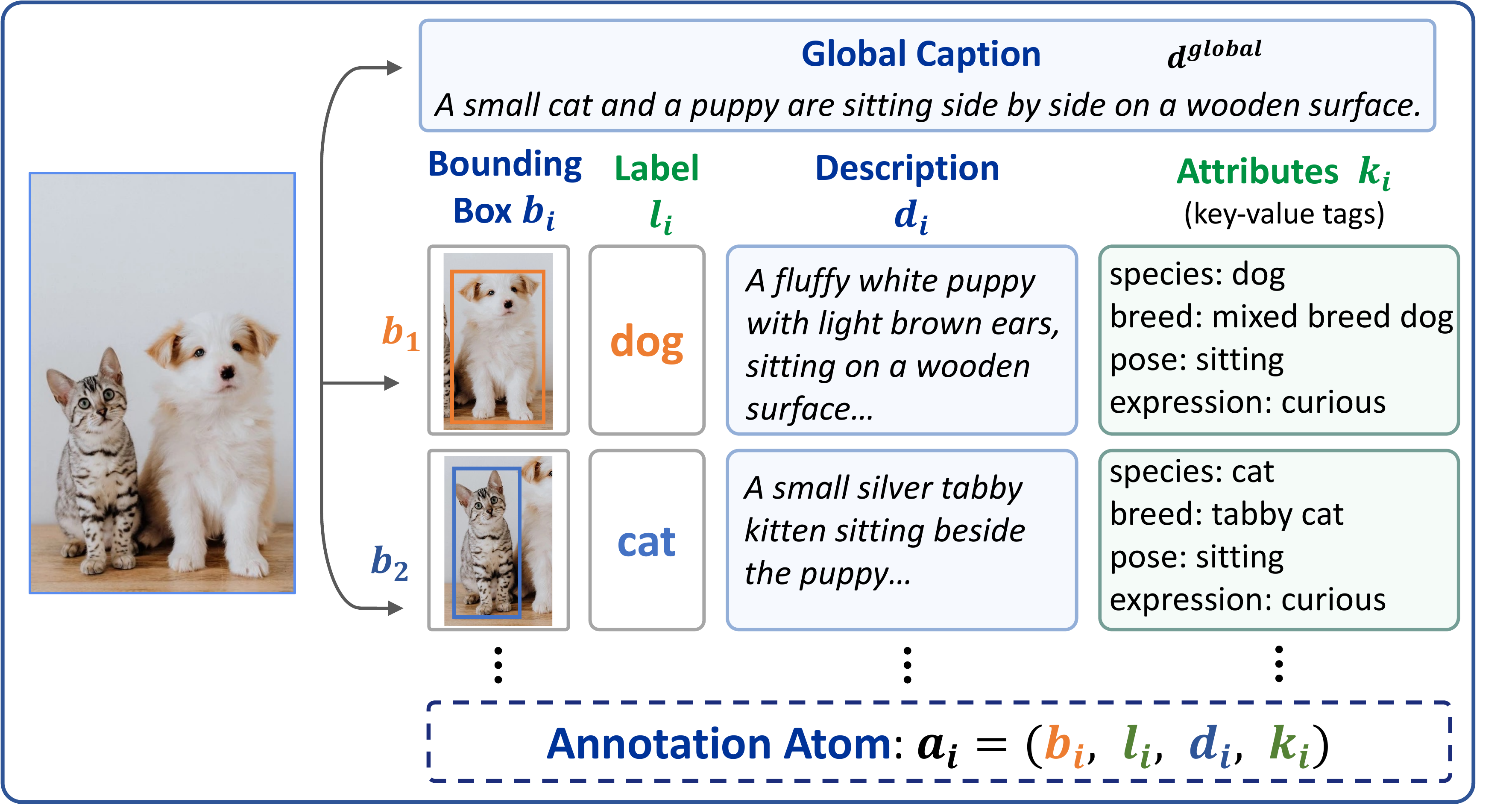}
    \caption{The unified annotation atom schema. Each bounding box is extended into an annotation atom with a categorical label, a free-form description, and structured key-value attributes defined by a task-specific declarative schema.}\label{fig:schema}
    \vspace{-1em}
\end{figure}

We define a unified annotation atom schema to capture the semantic richness required by VLMs, as shown in Figure~\ref{fig:schema}. 
For each image $I$, the schema supports the following annotations. An optional image-level caption $d$ describes the overall image. A set of bounding boxes $\{b_i\}_{i=1}^{N}$, each associated with a categorical label $\ell_i \in \{1, \dots, C\}$ from a task-specific label set. A per-box description $d_i$ providing a free-form textual description of the object, enabling open-vocabulary semantics beyond closed-set categories. Structured attributes $\mathbf{k}_i = \{(k_j, v_j)\}$, where the task administrators may define a declarative attribute schema specifying domain-specific properties for each label class. The annotation atom $a_i = (b_i, \ell_i, d_i, \mathbf{k}_i)$ unifies spatial localization, categorical identity, natural language semantics, and structured domain knowledge into a single representation. 

\subsection{On-Policy Annotation Loop}\label{sec:workflow}

To reduce human effort and maintain annotation consistency, we introduce an On-Policy Annotation Loop. As illustrated in Figure~\ref{fig:framework}(a), dataset construction proceeds iteratively over successive rounds. 
Each round involves two assistant models. The pre-annotation model $\pi_{k-1}$ comprises an object detector (e.g., YOLO) and a VLM. The annotation verifier $\phi_{k-1}$ performs quality control. At round $k$, a fresh batch of unlabeled images $\mathcal{B}_k$ is processed through the following stages, as shown in Figure~\ref{fig:framework}(a).

\noindent\textbf{Model-Assisted Pre-Annotation.}
Given the unlabeled batch $\mathcal{B}_k$, the policy $\pi_{k-1}$ automatically generates candidate annotation atoms. The object detector first proposes spatial bounding boxes and categories $\{(b_i, l_i)\}$. The VLM then fills in textual descriptions and structured attributes $\{(d_i, k_i)\}$ for each detected region.

\noindent\textbf{Human Review and Correction.}
A human annotator reviews the model's predictions, correcting erroneous boxes, labels, descriptions, and attributes. The human-corrected annotations are used to update $\mathcal{B}_k$. 

\noindent\textbf{Quality Control.}
The Bayesian Annotation Verifier $\phi_{k-1}$ scans annotations in $\mathcal{B}_k$ for spatial and categorical defects, flagging instances with high posterior uncertainty. A binary quality variable $z \in \{0, 1\}$ indicates whether an annotation is defective ($z=1$) or acceptable ($z=0$). Instances with high posterior uncertainty are flagged and routed back for a second human review within the current round.

\noindent\textbf{On-Policy Model Update.} 
After quality control, we merge $\mathcal{B}_k$ with the existing dataset to form the updated training set $\mathcal{D}_k = \mathcal{D}_{k-1} \cup \mathcal{B}_k$. All models are updated jointly on the full accumulated dataset: $\pi_k,\ \phi_k \leftarrow \text{train}(\mathcal{D}_k)$. The updated policy $\pi_k$ immediately supersedes $\pi_{k-1}$ for the next round of pre-annotation, while $\phi_k$ replaces $\phi_{k-1}$ as the verifier for round $k{+}1$.

\subsection{Bayesian Annotation Verifier}\label{sec:Bayesian}

The BAV ($\phi_k$) is a per-box binary classifier that estimates the posterior defect probability $p(z=1 \mid I, b, \ell)$ for each annotation.
It comprises three components: (1) a frozen vision backbone with a feature pyramid, (2) a geometric feature encoder, and (3) a Bayesian classification head.
Given an image $I$, a candidate box $b$, and a category index $\ell$, the backbone extracts a local RoI feature $\mathbf{v}_{\mathrm{box}}$ from the assigned pyramid level via RoI Align, together with a global context vector $\mathbf{v}_{\mathrm{global}}$ from the deepest level.
A geometric encoder maps the normalized box center, size, and aspect ratio to $\mathbf{v}_{\mathrm{geo}}$.
The three vectors are concatenated into $\mathbf{z} = [\mathbf{v}_{\mathrm{box}} \,\|\, \mathbf{v}_{\mathrm{global}} \,\|\, \mathbf{v}_{\mathrm{geo}}]$ and fed into a two-layer Bayesian linear head~\citep{blundell2015weight}, which maintains a variational posterior $q(W) = \mathcal{N}(\mu_W, \sigma_W^2)$ over its weights. 
The head outputs $C$-dimensional logits; the logit at the annotated category $\ell$ yields the defect probability:
$$
    r_\theta(I, b, \ell) = \sigma\bigl(g_\theta(I, b)_\ell\bigr) \approx p_\theta(z = 1 \mid I, b, \ell)
$$
Further architectural details are provided in Appendix~\ref{appendix:BAV}.

We construct BAV training data via error injection without manual quality labels. For each ground-truth annotation, we produce three samples: an acceptable sample ($z=0$) by retaining the original box or applying a small jitter with IoU $\geq 0.75$, a spatially corrupted sample ($z=1$) by displacing the box until IoU $\leq 0.4$, and a category-corrupted sample ($z=1$) by replacing the label with a randomly selected incorrect class.

\paragraph{Bayesian inference.} We apply MC Dropout \citep{gal2016dropout} with $T$ forward passes to obtain risk samples $\{r_\theta^{(t)}\}_{t=1}^{T}$, from which we compute two uncertainty metrics. Predictive entropy (total uncertainty):
\begin{align}
    \mathbb{H}[z \mid I, b, \ell] = -\bar{r}\log\bar{r} - (1-\bar{r})\log(1-\bar{r}), \label{eq:entropy}
\end{align}
where $\bar{r} = \frac{1}{T}\sum_{t=1}^{T} r_\theta^{(t)}$. And mutual information (epistemic uncertainty):
\begin{align}
    \mathbb{I}[z;\theta \mid I, b, \ell] &= \mathbb{H}[z \mid I, b, \ell] \nonumber\\
    &\quad - \frac{1}{T}\sum_{t=1}^{T}\mathbb{H}[z \mid I, b, \ell, \theta^{(t)}] \label{eq:mutual_info}
\end{align}
An annotation is flagged for re-review if $\bar{r} > \tau_r$ or $\mathbb{I} > \tau_I$, where both thresholds are task-specific.

\subsection{Multi-Task Data Synthesis}
A core bottleneck in scaling VLM training is the fragmented nature of data curation: introducing a new reasoning task conventionally requires re-annotating the entire corpus. 
To overcome this, our framework completely decouples data diversity from annotation cost through a template-driven multi-task data synthesis pipeline as shown in Figure~\ref{fig:framework}(b).
At its foundation, the unified annotation atom $a_i = (b_i, \ell_i, d_i, \mathbf{k}_i)$ provides a rich, multi-dimensional representation. This static data is dynamically unrolled into diverse conversational reasoning pairs via a three-layer abstraction. 
By simply applying different templates over the exact same set of underlying annotation atoms, a single labeled image instantly yields a rich spectrum of training tasks, e.g., attribute lookup, multi-hop path reasoning, and spatial inference. 
This paradigm ultimately maximizes the marginal supervision of the initial annotation, enabling the expansion of VLM capabilities without requiring any task-specific re-labeling.

\section{Experiment}

\begin{figure*}[ht]
    \centering
    \includegraphics[width=\linewidth]{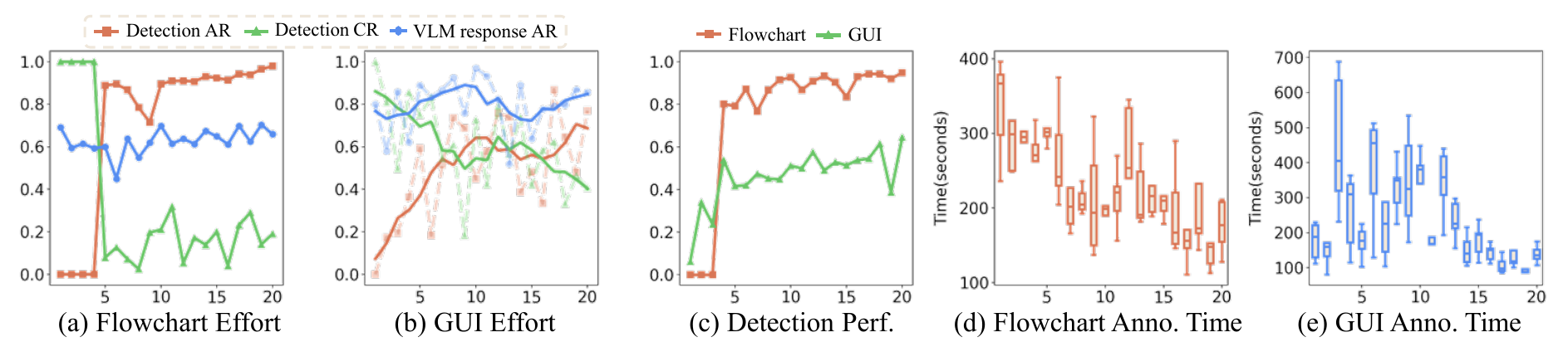}
    \caption{On-policy annotation efficiency (RQ1). (a)--(b): accept rate and completion rate per round for flowchart and GUI. (c): detection performance vs.\ cumulative annotated images. (d)--(e): annotation time per image across rounds. As annotation rounds progress, the accept rate rises, the completion rate and per-image annotation time both decline, and detector accuracy improves steadily, confirming that the on-policy loop progressively transfers labeling effort from human annotators to the model.}
    \label{fig:rq1}
\end{figure*}

Our experiments are organized around three research questions:

\noindent\textbullet\ \textbf{RQ1.} To what extent does the on-policy annotation loop reduce human labeling effort?

\noindent\textbullet\ \textbf{RQ2.} Can the Bayesian Annotation Verifier reliably detect annotation errors?

\noindent\textbullet\ \textbf{RQ3.} Can annotation atoms be reused to synthesize effective training data for diverse VLM tasks?

\subsection{Experimental Settings} 

We evaluate ScreenAnnotator on two scenarios: flowchart and mobile GUI screenshot. Flowcharts require relational annotation with directed edges carrying explicit source/target references. GUI screenshots demand open-vocabulary grounding over visually diverse, context-dependent UI elements. For the flowchart scenario, we use 100 images from the Flowchart-QA dataset\footnote{\url{https://huggingface.co/datasets/Kingsoft-LLM/QZhou-Flowchart-QA}} for training and 40 for evaluation. For the GUI scenario, we collect 120 screenshots via script-driven exploration and manual capture on mobile phones, with 20 held out for evaluation. 

In the on-policy annotation loop, we employ a YOLOv11-based detector and a Qwen3-VL-4B-Instruct\citep{bai2025qwen3} VLM as the initial models, which are jointly updated after each round.
Specifically, we use YOLOv11n\citep{yolo11_ultralytics} for the flowchart scenario and YOLOv11s for the GUI scenario. All experiments are conducted using a single NVIDIA L40 GPU.

\subsection{On-Policy Annotation Efficiency (RQ1)}

The on-policy loop runs in rounds of 5 images. We quantify labeling effort along three dimensions.
Accept Rate (AR) is the fraction of model-proposed annotations that human annotators accept without any modification in a given round. A higher accept rate indicates that less human intervention is required.
Completion Rate (CR) is the fraction of annotation atoms that humans must complete because the model failed to correctly identify them.
Annotation Time measures the wall-clock time an annotator spends on a single image, averaged over all images in a round.
Figure~\ref{fig:rq1} summarizes the results for these three efficiency metrics and task-specific performance (Appendix~\ref{appendix:metrics}).

Figure~\ref{fig:rq1} (a) and Figure~\ref{fig:rq1} (b) report the annotation effort per round.
In the flowchart scenario (Figure~\ref{fig:rq1} (a)), the detector confidence remains below the acceptance threshold of 0.7 throughout the first~5 rounds, reflecting the inevitable cold-start behavior of task-specific training. All images in this period therefore require full manual completion.
After round~5, the detector begins to produce increasingly accurate proposals. The accept rate rises steadily to nearly 100\%, while the completion rate oscillates around 20\%.
The VLM accept rate exhibits a slow upward trend around 60\%.
In Figure~\ref{fig:rq1} (b), the GUI scenario follows a similar trajectory. The accept rate increases from 0\% in round~1 to 77\% by round~20, while the completion rate drops from 100\% to 40.6\%. Meanwhile, the VLM accept rate fluctuates around 80\%.
Both curves confirm that successive retraining cycles progressively shift the annotation workload from human annotators to the model.
Figure~\ref{fig:rq1}~(c) plots the detection performance against annotation rounds.
In both cases, as the amount of labeled data increases, the accuracy of the detectors improves steadily. Especially during the cold start phase, the performance of the detectors significantly improves. The accuracy of the flowchart detector ultimately reaches 94.9\%, while the accuracy of the GUI detector reaches 64.4\%.

Figure~\ref{fig:rq1} (d) and Figure~\ref{fig:rq1} (e) present box plots of per-image annotation time across rounds.
Both the median time and the within-batch variance decrease substantially as the loop progresses.
The reduction is consistent with the rising accept rate. As the model produces more accurate proposals, annotators spend less time on corrections and can approve suggestions with minimal modification.
Taken together, these results establish that the on-policy loop reduces labeling cost along all three dimensions simultaneously.

\subsection{Bayesian Annotation Verifier (RQ2)}

We measure BAV's error-prioritization ability with the lift metric. Lift measures how effectively the ranked quality scores place true errors within the top-$k$ fraction compared with random inspection. Note that a value of 1.0 denotes no gain. The formal definition of lift is denoted in Appendix~\ref{appendix:lift}.
BAV is trained in a fully self-supervised manner via controlled perturbation of accepted annotations, requiring no manual quality labels.
During inference, $T{=}30$ MC-Dropout passes yield a per-box quality score with the threshold $\tau{=}0.5$. Annotations are then ranked by this score, and we report lift at budget levels of 1\%, 5\%, and 10\% across on-policy rounds in both scenarios.

\begin{figure}[t]
    \centering
    \includegraphics[width=\linewidth]{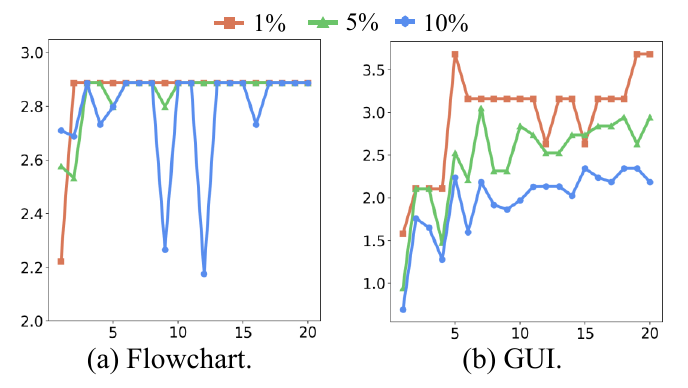}
    \caption{Lift score of BAV at budget levels 1\%, 5\%, and 10\%. A higher lift score means that true errors are concentrated earlier in the ranked list, so a smaller inspection budget suffices to capture them.}
    \label{fig:rq2}
\end{figure}

Figure~\ref{fig:rq2} plots the lift curves across on-policy rounds for both scenarios.
In the flowchart scenario (Figure~\ref{fig:rq2} (a)), lift@1\% reaches 2.89 by round~3 and remains at this level throughout the remaining rounds, indicating that the top 1\% of flagged boxes captures roughly 2.9$\times$ more true errors than random sampling.
Lift@5\% and lift@10\% converge to the same plateau, both stabilizing at 2.89.
The rapid saturation reflects the relatively uniform error distribution in flowchart annotations. Once sufficient perturbation examples are available, the verifier's ranking quality plateaus.
In the GUI scenario (Figure~\ref{fig:rq2} (b)), lift@1\% climbs from 1.58 in round~1 to 3.69 by round~5, before stabilizing around 3.16--3.69 in later rounds. Meanwhile, lift@5\% and lift@10\% exhibit a steady upward trend, reaching 2.95 and 2.35, respectively, by round~20.
The higher peak lift in the GUI scenario suggests that annotation errors are more heterogeneous, allowing the uncertainty-based ranking to achieve stronger separation.
Across both scenarios, the lift values increase with the number of on-policy rounds and remain well above 1.0, confirming that the BAV's ranking quality improves as the training pool grows.

These results demonstrate that BAV can effectively prioritize annotation errors for re-review without any human quality supervision.
Even with only a 10\% inspection budget, the verifier surfaces a large fraction of true errors near the top of the ranked list, substantially reducing manual quality-assurance effort.

\subsection{Template-Driven VLM Data Synthesis (RQ3)}\label{sec:rq3}

Based on the flowchart schema, we instantiate chat templates for 12 task types (Appendix~\ref{appendix:templates}) and synthesize QA pairs directly from the annotation atoms. The resulting tasks span attribute lookup, relational reasoning, path traversal, counting, and spatial inference.
We fine-tune Qwen3-VL-8B-Instruct on the synthesized data and evaluate it on a held-out set of 20 flowchart images comprising 240 manually verified QA pairs (one per task type per image). Each response is scored by human judges using a structured rubric that decomposes answers into textual and spatial components with task-specific weights (Appendix~\ref{appendix:scoring}).

\begin{figure}[t]
    \centering
    \includegraphics[width=0.85\columnwidth]{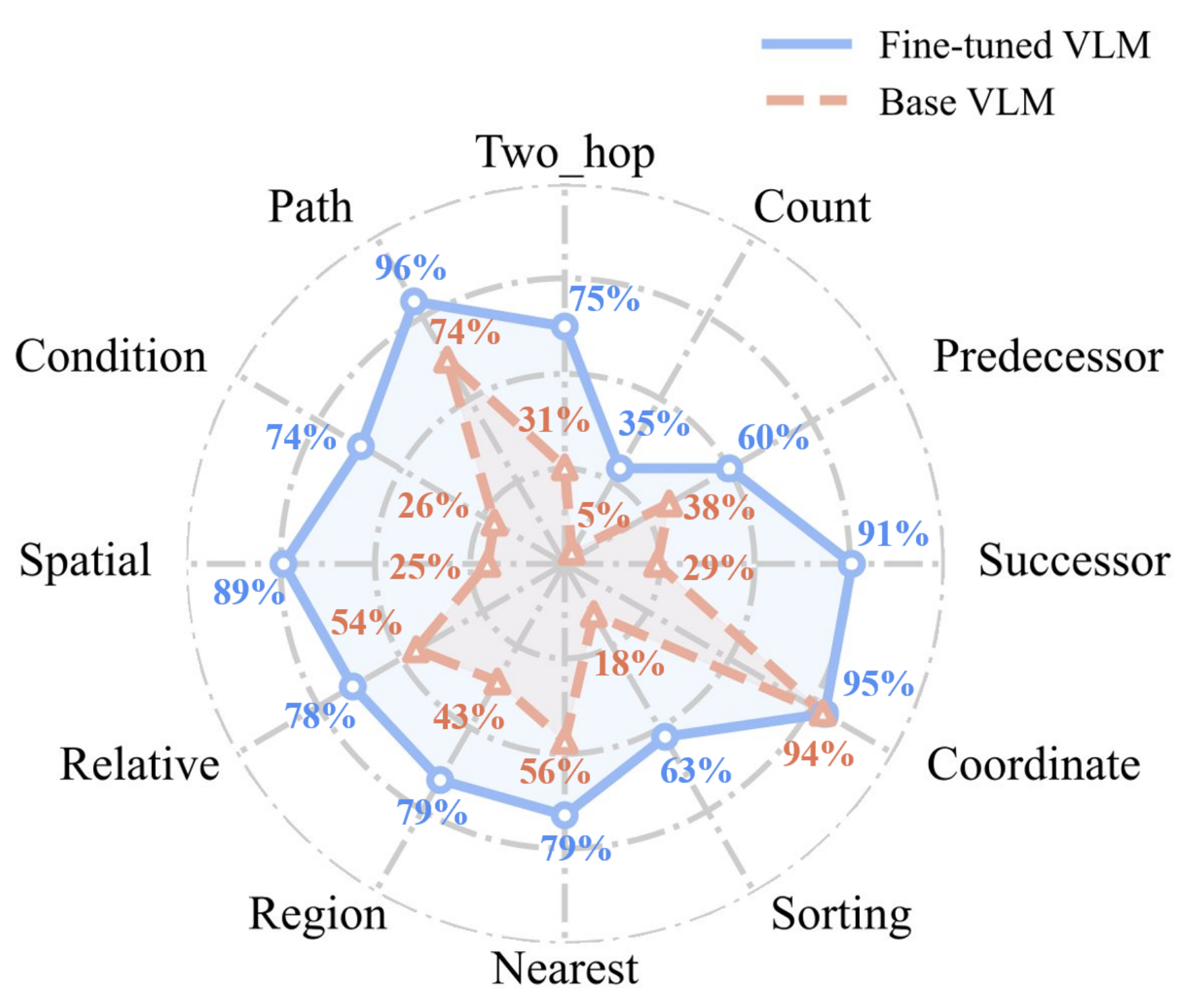}
    \caption{Per-task accuracy on the flowchart test set before and after fine-tuning with synthesized QA data (RQ3).}
    \label{fig:rq3_radar}
    \vspace{-1em}
\end{figure}

\begin{figure*}[ht]
    \centering
    \includegraphics[width=\linewidth]{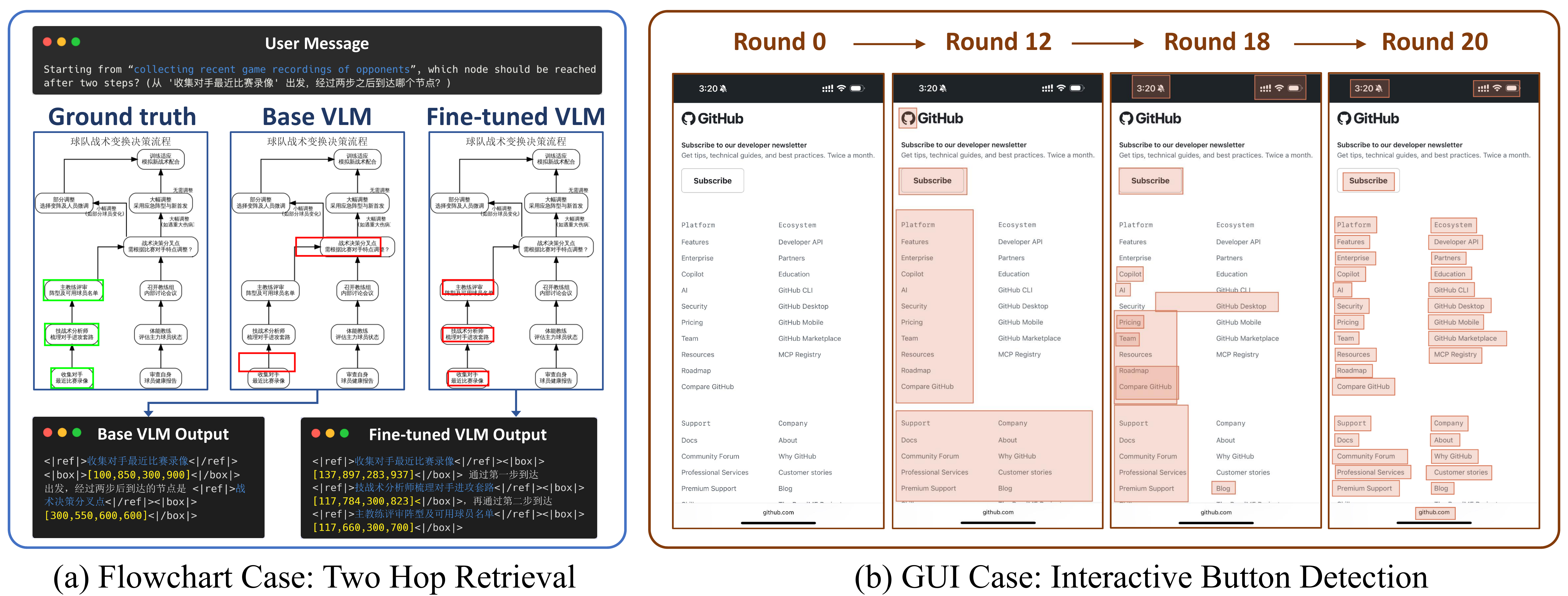}
    \caption{Qualitative examples. (a) Two-hop retrieval on a flowchart: the base VLM skips intermediate reasoning and directly produces an incorrect destination, whereas the fine-tuned model precisely identifies both the intermediate node and the final destination. (b) Interactive button detection on a GUI screenshot: predictions from the detector at round~0 and at round~20, showing substantially improved recall and localization accuracy after on-policy retraining.}
    \label{fig:case}
\end{figure*}

Figure~\ref{fig:rq3_radar} presents a comparison of per-task accuracy between the base VLM and the fine-tuned model. Overall, the fine-tuned model achieves an average accuracy of 76.1\%, substantially outperforming the base model’s 41.0\%, corresponding to an absolute improvement of 35.1\%. Performance gains are consistently observed across all 12 task types, with the most pronounced improvements appearing in tasks that require structured understanding of flowchart topology. Successor identification jumps from 29.3\% to 90.6\%, while condition reasoning rises from 25.9\% to 74.4\%. These gains indicate that the synthesized relational supervision enables the model to resolve directional edge semantics that the base model largely fails to capture.
Two-hop reasoning and sorting, both of which demand chaining multiple inference steps, improve by 44.4\% and 44.8\% in absolute terms.
Spatial inference exhibits the single largest absolute gain, rising from 24.5\% to 89.0\%, confirming that the bounding-box grounding tokens embedded in the synthesized data transfer effectively to coordinate-level reasoning.
Region and relative position tasks also benefit substantially, increasing by 36.0\% and 23.3\% respectively.
Path traversal reaches 95.5\% (up from 74.3\%), reflecting the model's improved ability to trace multi-node sequences once explicit path annotations are available.
Counting remains the most challenging task at 34.8\%, yet still represents a 30.3\% absolute improvement over the near-zero baseline of 4.5\%. This result suggests that enumeration over dense, visually similar nodes poses difficulties beyond what template-level supervision alone can resolve.
Coordinate localization, where the base model already scores 94.3\%, shows only a marginal gain to 95.0\%, indicating a performance ceiling for this relatively simple grounding task.
Notably, all training data originate from the same set of annotation atoms produced by the on-policy loop.
Adding a new task type requires only defining an additional chat template, with no re-annotation of images.

\subsection{Case Study}

Figure~\ref{fig:case} provides two qualitative examples.
Figure~\ref{fig:case}~(a) shows that template-driven fine-tuning enables the model to perform structured multi-hop reasoning over flowchart topology, which is a capability absent in the base VLM.
Figure~\ref{fig:case}~(b) confirms that on-policy retraining substantially improves detection coverage and localization precision across annotation rounds, which is consistent with the efficiency trends reported in Figure~\ref{fig:rq1}.

\section{Conclusion}
Existing annotation infrastructures struggle to meet the intricate data demands of VLMs for complex reasoning tasks. To address this critical limitation, we introduced \textsc{ScreenAnnotator} in this paper.
Specifically, it defines a unified annotation atom schema that seamlessly binds spatial coordinates, open-vocabulary descriptions, and structured attributes. Furthermore, an on-policy annotation loop ensures continuous model-human co-evolution and self-supervised quality control. Finally, a template-driven multi-task synthesis procedure dynamically unrolls these static atoms into highly diverse conversational reasoning tasks.
We validated our framework on two annotation-intensive scenarios: flowchart parsing and GUI screenshot understanding. 
Experimental results demonstrate that our tool substantially improves annotation efficiency and effectively detects annotation errors. More importantly, datasets constructed with our tool significantly boost VLM performance on complex visual reasoning tasks, validating the practical value and potential of our approach.
Overall, \textsc{ScreenAnnotator} provides a highly scalable and easy-to-use solution for curating datasets that satisfy advanced VLM reasoning needs.

\section*{Limitations}

\noindent\textbf{Scope of scenarios.} Our experiments focus on flowchart parsing and GUI screenshot understanding, two domains that share dense objects, explicit spatial relations, and a semi-open category space. These properties align well with the compositional annotation atom and the on-policy loop. Scaling to larger annotation volumes and extending the framework to domains with different annotation semantics, such as medical imaging and remote sensing, is a promising direction for future work.

\noindent\textbf{Sample selection and uncertainty granularity.} The present work concentrates on pre-annotation, human correction, and iterative retraining. Images currently enter the pipeline in a fixed order, and training batches are sampled uniformly. Incorporating active sample selection strategies could further improve annotation efficiency. Similarly, the Bayesian Annotation Verifier currently operates at the bounding-box level. Extending uncertainty estimation to free-form descriptions and structured attributes remains an interesting avenue to explore.

\section*{Ethical Considerations}

All images used in our experiments are either sourced from publicly available datasets with permissive licenses or collected by the authors themselves, and all usage is consistent with the original intended purposes and access conditions. GUI screenshots were captured from the authors' own devices and manually inspected to ensure no personally identifiable information or offensive content is present. Human annotators involved in the on-policy loop were fully informed about the purpose of the annotation task and compensated with reasonable remuneration. All artifacts we release are intended solely for academic research. We do not foresee direct negative societal impacts from this work, as the framework is intended to improve data quality and reduce repetitive manual labor rather than to automate decisions affecting individuals.

\bibliography{custom}

\newpage
\newpage
\appendix
\section{Appendix}

\subsection{Bayesian Annotation Verifier Architecture}\label{appendix:BAV}

This section provides the full architectural and training details of the Bayesian Annotation Verifier (BAV) introduced in \S\ref{sec:Bayesian}.

\paragraph{Overview.}
BAV is a per-box binary classifier that estimates the probability of an annotation being defective, conditioned on the image, the bounding box coordinates, and the annotated category label.
The model comprises three components: (1) a frozen vision backbone with a feature pyramid, (2) a geometric feature encoder, and (3) a Bayesian classification head.
Given an image $I$, a candidate box $b = (x_1, y_1, x_2, y_2)$, and a category index $\ell$, the model outputs:
\begin{equation}
    r_\theta(I, b, \ell) = \sigma\!\bigl(g_\theta(I, b)_\ell\bigr) \approx p_\theta(z = 1 \mid I, b, \ell)
\end{equation}
where $\theta$ denotes a sample from the variational posterior over the network weights.

\paragraph{Vision backbone and RoI features.}
We adopt a SAM3 vision encoder as the backbone, which produces a four-level feature pyramid with strides $\{3.5, 7.0, 14.0, 28.0\}$ and a uniform channel dimension of $C_{\mathrm{fpn}}{=}256$.
Input images are resized to $1008 \times 1008$.
Each candidate box is assigned to a pyramid level based on its area:
\begin{equation}
    k = \left\lfloor k_0 + \log_2\!\left(\frac{\sqrt{(x_2{-}x_1)(y_2{-}y_1)}}{s_0}\right) \right\rfloor
\end{equation}
with canonical scale $s_0{=}224$ and canonical level $k_0{=}2$, clamped to $[0, 3]$.
RoI Align~\citep{he2017mask} is applied at the assigned level with a $7 \times 7$ output resolution, yielding a local feature map that is flattened and linearly projected to $\mathbf{v}_{\mathrm{box}} \in \mathbb{R}^{256}$.
A global context vector $\mathbf{v}_{\mathrm{global}} \in \mathbb{R}^{256}$ is obtained by average-pooling the deepest pyramid level over its spatial dimensions.

\paragraph{Geometric features.}
A five-dimensional descriptor is computed for each box:
\begin{equation}
    \mathbf{g} = \left[\frac{c_x}{W},\; \frac{c_y}{H},\; \frac{w}{W},\; \frac{h}{H},\; \frac{w}{h+\epsilon}\right]
\end{equation}
encoding the normalized center, normalized width and height, and the aspect ratio.
A single linear layer with ReLU activation maps $\mathbf{g}$ to $\mathbf{v}_{\mathrm{geo}} \in \mathbb{R}^{32}$.

\paragraph{Bayesian classification head.}
The three feature vectors are concatenated into $\mathbf{z} = [\mathbf{v}_{\mathrm{box}} \,\|\, \mathbf{v}_{\mathrm{global}} \,\|\, \mathbf{v}_{\mathrm{geo}}] \in \mathbb{R}^{544}$ and passed through a two-layer network with Bayesian linear layers~\citep{blundell2015weight}.
Each Bayesian layer maintains a variational posterior $q(W) = \mathcal{N}(\mu_W, \sigma_W^2)$ over its weights, where $\sigma_W = \mathrm{softplus}(\rho_W)$.
During each forward pass, weights are sampled via the reparameterization trick:
\begin{equation}
    W = \mu_W + \sigma_W \odot \varepsilon, \quad \varepsilon \sim \mathcal{N}(0, I)
\end{equation}
The prior is an isotropic Gaussian $p(W) = \mathcal{N}(0, \sigma_{\mathrm{prior}}^2 I)$ with $\sigma_{\mathrm{prior}}{=}1.0$.
The output layer produces $C$-dimensional logits, and the logit corresponding to the annotated category $\ell$ is used for binary classification.

\paragraph{Training objective.}
The model is trained by minimizing the negative evidence lower bound (ELBO):
\begin{equation}
    \mathcal{L} = \mathrm{BCE}_{\mathrm{weighted}}(\hat{z}, z) + \lambda_{\mathrm{KL}} \sum_{l} \mathrm{KL}\!\left[q(\theta_l) \,\|\, p(\theta_l)\right]
\end{equation}
The KL divergence for each Bayesian layer is computed in closed form between two Gaussians.
To handle the imbalanced ratio between positive (defective) and negative (acceptable) samples, the binary cross-entropy term applies a positive-class weight $w_{\mathrm{pos}} = N_{\mathrm{neg}} / N_{\mathrm{pos}}$.
The KL weight is set to $\lambda_{\mathrm{KL}} = 1/N$ to balance its magnitude with the likelihood term.

\paragraph{Self-supervised training data.}
BAV requires no manually labeled quality annotations.
Training samples are generated by programmatic perturbation of ground-truth boxes.
For each annotation, three samples are produced: (1) an acceptable sample ($z{=}0$), obtained by keeping the original box or applying a small jitter with IoU $\geq 0.75$; (2) a spatially corrupted sample ($z{=}1$), produced by displacing the box until IoU $\leq 0.4$; and (3) a category-corrupted sample ($z{=}1$), where the box coordinates are preserved but the label is replaced with a randomly selected incorrect class.

\paragraph{Optimization.}
We use AdamW with a base learning rate $\eta$ and a reduced rate $\eta \times 0.1$ for the backbone parameters.
The backbone is frozen for the first 10 epochs and then unfrozen for joint fine-tuning.

\paragraph{Inference.}
At inference time, the Bayesian layers remain stochastic while the backbone batch normalization statistics are fixed.
We perform $T{=}30$ Monte Carlo forward passes to obtain risk samples $\{r_\theta^{(t)}\}_{t=1}^{T}$.
The mean risk score $\bar{r} = \frac{1}{T}\sum_t r_\theta^{(t)}$ serves as the defect score.
Predictive entropy and mutual information are computed as described in \S\ref{sec:Bayesian}.
An annotation is flagged when $\bar{r} \geq \tau$ with default threshold $\tau{=}0.5$.

\subsection{Task Evaluation Metrics}\label{appendix:metrics}

We evaluate two aspects of annotation quality separately: bounding box detection and node text recognition.

\textbf{Bounding box detection.}
We report the Detection F1 score, which quantifies whether each element is both localized and categorized correctly.
Given a set of ground-truth boxes and predicted boxes, predictions are matched to ground truths greedily in descending order of confidence using IoU as the matching criterion.
A prediction is counted as a true positive (TP) only if its IoU with the matched ground-truth box meets or exceeds the threshold \emph{and} its category label is correct; a spatial match with an incorrect label is counted as a false positive (FP).
Unmatched ground-truth boxes are false negatives (FN).
Each ground-truth box is matched to at most one prediction, and each prediction to at most one ground-truth box.
Detection F1 is computed as $\text{F1} = 2\text{TP} / (2\text{TP} + \text{FP} + \text{FN})$ at five IoU thresholds $\{0.55, 0.65, 0.75, 0.85, 0.95\}$, and we report the mean across thresholds.

\textbf{Node text recognition.}
For flowchart nodes containing text, transcription quality is evaluated with two complementary metrics: Exact Match Rate (EMR) and 1-NED.
EMR measures the fraction of nodes whose predicted text is identical to the ground-truth string after normalization (stripping whitespace, unifying punctuation and full-/half-width characters, and lowercasing for Latin text).
The 1-NED score is defined as $1 - \text{EditDistance}(\hat{y}, y) / \max(|\hat{y}|, |y|)$, where EditDistance denotes the Levenshtein distance; it captures partial correctness and ranges from 0 (completely dissimilar) to 1 (exact match).
Both metrics are averaged over all annotated text nodes in the test set.

\subsection{Lift Metric}\label{appendix:lift}

We use the lift metric to evaluate how effectively the BAV quality score prioritizes erroneous annotations for human re-review.
Given a set of $N$ candidate annotations, each annotation is assigned a quality score (the estimated probability of being erroneous).
All candidates are sorted by this score in descending order.
For a budget ratio $\alpha \in (0, 1]$, we select the top-$k$ candidates where $k = \lceil N \cdot \alpha \rceil$, clamped to $[1, N]$.

Let $P$ denote the total number of truly erroneous annotations in the full candidate set, and let $P_k$ denote the number of truly erroneous annotations within the selected top-$k$ subset.
The lift at budget $\alpha$ is defined as:
\begin{equation}
    \text{lift}@\alpha = \frac{P_k / P}{k / N}
\end{equation}
The numerator $P_k / P$ is the recall of erroneous annotations within the inspected subset.
The denominator $k / N$ is the fraction of annotations actually inspected.
Lift can equivalently be expressed as the ratio of the precision within the top-$k$ to the global error rate:
\begin{equation}
    \text{lift}@\alpha = \frac{P_k / k}{P / N}
\end{equation}
A lift of 1.0 corresponds to random ranking, where the inspected subset contains errors at the same rate as the full set.
Values greater than 1.0 indicate that the ranking concentrates errors toward the top, enabling annotators to catch more mistakes with less review effort.

Note that the effective budget fraction $k/N$ may differ slightly from the nominal ratio $\alpha$ due to the ceiling operation, particularly when $N$ is small.
For example, with $N{=}23$ and $\alpha{=}0.1$, we have $k{=}3$ and the actual budget fraction is $3/23 \approx 0.130$.
The lift computation always uses the realized fraction $k/N$.

\subsection{Human Scoring Rubric for Multi-Task Evaluation in Flowchart Scenario}\label{appendix:scoring}

Each synthesized QA sample is scored by a human judge on a $[0, 1]$ scale.
The rubric is designed so that every instance of the same task type is graded under a uniform standard.
Scores are decomposed into a textual component and a spatial (bounding box) component, with task-specific weights summarized in Table~\ref{tab:scoring_weights}.

\paragraph{Textual answers.}
A fully correct textual response receives full credit.
Minor errors affecting only one or two characters receive approximately 60\% credit.
Responses that are substantially wrong receive zero credit.

\paragraph{Bounding boxes.}
Spatial predictions are graded on a five-level scale: 100\% for a precisely localized box, 80\% for a box that covers the target with slight misalignment, 50\% for a box whose intent is recognizable but whose localization is poor, 20\% for a box that is far from the target yet shows a directional tendency, and 0\% for a missing box, an unparseable output, or a degenerate full-image box.

\paragraph{Task-specific weighting.}
Different task types emphasize different answer components.
For tasks that require identifying target nodes (e.g., Successor, Predecessor, Condition), the score is a weighted sum of the query-node localization and the answer-node localization.
For Count, 30\% of the score is assigned to the numerical answer and 70\% to the quality of the enumerated boxes.
For Path and Two-hop, credit is allocated across each node along the expected trajectory.
For Spatial, Relative, Nearest, and Region, the textual direction or identity answer and the box quality each contribute 50\%.
For Sorting, 80\% of the score depends on the correctness and ordering of the answer, with the remaining 20\% on box quality; completely random boxes receive zero.
For Coordinate, text and box quality are weighted equally.
When multiple valid answers exist, any correct alternative is accepted.

\begin{table}[t]
\centering
\small
\caption{Per-task scoring weight decomposition. ``Text'' denotes the textual answer component and ``Box'' the spatial localization component.}
\label{tab:scoring_weights}
\begin{tabular}{lcc}
\toprule
Task Type & Text Weight & Box Weight \\
\midrule
Successor / Predecessor & 0.30 & 0.70 \\
Count & 0.30 & 0.70 \\
Two-hop & \multicolumn{2}{c}{0.2 / 0.4 / 0.4 (per node)} \\
Path & \multicolumn{2}{c}{0.6 (path) + box quality} \\
Condition & 0.20 + 0.30 & 0.50 \\
Spatial / Relative / & \multirow{2}{*}{0.50} & \multirow{2}{*}{0.50} \\
Nearest / Region & & \\
Sorting & 0.80 & 0.20 \\
Coordinate & 0.50 & 0.50 \\
\bottomrule
\end{tabular}
\end{table}

\lstdefinestyle{tmpl}{
  basicstyle=\ttfamily\scriptsize,
  frame=single,
  breaklines=true,
  columns=fullflexible,
  aboveskip=4pt,
  belowskip=4pt,
  xleftmargin=2pt,
  xrightmargin=2pt,
}

\subsection{VLM Prompts for Attribute Extraction}\label{appendix:vlm-prompts}

Our tool supports generating distinct attributes for different annotation atoms—for example, source/target nodes and condition text for flowchart edges, and functional descriptions for GUI elements. These attributes are produced by specially designed VLM prompt templates. Below we present the complete prompts designed for two representative tasks:

\newpage
\paragraph{Flowchart edge attribute extraction.}\mbox{}
\begin{lstlisting}[style=tmpl]
You are a flowchart edge relationship recognition assistant. The user will provide you with a complete flowchart image. 
The target edge is specified by a bbox text description, where the bbox coordinates are normalized to a 0-1000 coordinate system based on the full image width and height, with the format [x_min, y_min, x_max, y_max]. This bbox is guaranteed to correspond to an edge in the flowchart. 
You will also be given a list of neighboring candidate nodes for this bbox, where each node contains an id and text. 
Your task is to determine the source node, the target node, and any condition text or descriptive text associated with the edge. 
The values of from and to must strictly use ids from the candidate node list. If there is no text on the edge, output an empty string for text. 
Strictly output only a JSON object. Do not output markdown, explanations, or any additional text. The format must be: {"text":"...","from":"...","to":"..."}
\end{lstlisting}

\paragraph{GUI region description.}\mbox{}
\begin{lstlisting}[style=tmpl]
You are a professional GUI/image region description assistant. 
A complete screenshot will be provided below to help you understand the context. 
The target region is specified by a bbox text description, where the bbox coordinates are normalized to a 0-1000 coordinate system based on the full image width and height, with the format [x_min, y_min, x_max, y_max]. 
Please describe the function of the given local region using the most concise single word or short phrase possible, such as close, back, menu, etc. 
Do not output markdown, JSON, or any additional explanations.
\end{lstlisting}

\subsection{Chat Templates for Flowchart QA Synthesis}\label{appendix:templates}

This section lists the complete set of chat templates used to synthesize VLM training data from flowchart annotation atoms (\S\ref{sec:rq3}).
All tasks share a common system prompt.
Placeholders in curly braces are instantiated from the annotation graph at synthesis time.
All bounding-box coordinates are normalized to the $[0, 1000]$ range, required by Qwen-VL.

\paragraph{System prompt (shared across all tasks).}\mbox{}
\begin{lstlisting}[style=tmpl]
You are a flowchart understanding assistant. Answer the question based only on the image content. When referring to nodes, edges, or regions, use the specified format to output the corresponding text and coordinate reference whenever possible. All bbox coordinates are normalized to the 0-1000 range, for example: <|ref|>node text<|/ref|><|box|>[x1,y1,x2,y2]<|/box|>. Do not output explanations unrelated to the question.
\end{lstlisting}

\paragraph{1.\;Successor.}\mbox{}
\begin{lstlisting}[style=tmpl]
Q: In this flowchart, what is the next step after {node.text}?
A: The next step after {node.text} is <|ref|>{successors(node)[0].text}<|/ref|><|box|>{successors(node)[0].bbox}<|/box|>.
\end{lstlisting}

\paragraph{2.\;Predecessor.}\mbox{}
\begin{lstlisting}[style=tmpl]
Q: In this flowchart, which step directly connects to {node.text}?
A: The step that directly connects to {node.text} is <|ref|>{predecessors(node)[0].text}<|/ref|><|box|>{predecessors(node)[0].bbox}<|/box|>.
\end{lstlisting}

\paragraph{3.\;Count.} 
This task type has four variants: node count, edge count, out-degree, and in-degree.

\begin{lstlisting}[style=tmpl]
Q: How many nodes are there in this flowchart?
A: There are {len(nodes)} nodes in this flowchart <|ref|>{len(nodes)} nodes<|/ref|><|box|>[{nodes[0].bbox}, {nodes[1].bbox}, ..., {nodes[n].bbox}]<|/box|>.
\end{lstlisting}
\begin{lstlisting}[style=tmpl]
Q: How many connecting edges are there in this flowchart?
A: There are {len(edges)} edges in this flowchart <|ref|>{len(edges)} edges<|/ref|><|box|>[{edges[0].bbox}, {edges[1].bbox}, ..., {edges[m].bbox}]<|/box|>.
\end{lstlisting}
\begin{lstlisting}[style=tmpl]
Q: How many outgoing edges does {node.text} have?
A: <|ref|>{node.text}<|/ref|><|box|>{node.bbox}<|/box|> has {len(successors(node))} outgoing edges, connecting to {successors(node)[0].text}, {successors(node)[1].text}, ..., {successors(node)[k].text} <|ref|>{len(successors(node))} successor nodes<|/ref|><|box|>[{successors(node)[0].bbox}, {successors(node)[1].bbox}, ..., {successors(node)[k].bbox}]<|/box|>.
\end{lstlisting}
\begin{lstlisting}[style=tmpl]
Q: How many steps directly connect to {node.text}?
A: There are {len(predecessors(node))} steps that directly connect to <|ref|>{node.text}<|/ref|><|box|>{node.bbox}<|/box|>, namely {predecessors(node)[0].text}, {predecessors(node)[1].text}, ..., {predecessors(node)[k].text} <|ref|>{len(predecessors(node))} predecessor nodes<|/ref|><|box|>[{predecessors(node)[0].bbox}, {predecessors(node)[1].bbox}, ..., {predecessors(node)[k].bbox}]<|/box|>.
\end{lstlisting}

\paragraph{4.\;Two-hop.}\mbox{}
\begin{lstlisting}[style=tmpl]
Q: Starting from {path[0].text}, which node is reached after two steps?
A: Starting from <|ref|>{path[0].text}<|/ref|><|box|>{path[0].bbox}<|/box|>, the first step reaches <|ref|>{path[1].text}<|/ref|><|box|>{path[1].bbox}<|/box|>, and the second step reaches <|ref|>{path[2].text}<|/ref|><|box|>{path[2].bbox}<|/box|>.
\end{lstlisting}

\paragraph{5.\;Path.}\mbox{}
\begin{lstlisting}[style=tmpl]
Q: List the complete path from {path[0].text} to {path[-1].text}.
A: The complete path is: <|ref|>{path[0].text}<|/ref|><|box|>{path[0].bbox}<|/box|> -> <|ref|>{path[1].text}<|/ref|><|box|>{path[1].bbox}<|/box|> -> ... -> <|ref|>{path[-1].text}<|/ref|><|box|>{path[-1].bbox}<|/box|>.
\end{lstlisting}

\paragraph{6.\;Condition.}\mbox{}
\begin{lstlisting}[style=tmpl]
Q: When the decision result of {edge.from_node.text} is {edge.text}, where does the flow go?
A: When the result of <|ref|>{edge.from_node.text}<|/ref|><|box|>{edge.from_node.bbox}<|/box|> is <|ref|>{edge.text}<|/ref|><|box|>{edge.bbox}<|/box|>, the flow goes to <|ref|>{edge.to_node.text}<|/ref|><|box|>{edge.to_node.bbox}<|/box|>.
\end{lstlisting}

\paragraph{7.\;Spatial.}\mbox{}
\begin{lstlisting}[style=tmpl]
Q: Where is {node.text} located in the image?
A: <|ref|>{node.text}<|/ref|><|box|>{node.bbox}<|/box|> is located in the {region(node.bbox)} region of the image.
\end{lstlisting}

\paragraph{8.\;Relative.}\mbox{}
\begin{lstlisting}[style=tmpl]
Q: In which direction is {first_node.text} relative to {second_node.text}?
A: <|ref|>{first_node.text}<|/ref|><|box|>{first_node.bbox}<|/box|> is located to the {relative_pos(first_node.bbox, second_node.bbox)} of <|ref|>{second_node.text}<|/ref|><|box|>{second_node.bbox}<|/box|>.
\end{lstlisting}

\paragraph{9.\;Region.}\mbox{}
\begin{lstlisting}[style=tmpl]
Q: Which nodes are located in the {target_region} region of the image?
A: The nodes located in the {target_region} region are: {node_1.text}, {node_2.text}, ..., {node_k.text} <|ref|>{target_region} region nodes<|/ref|><|box|>[{node_1.bbox}, {node_2.bbox}, ..., {node_k.bbox}]<|/box|>.
\end{lstlisting}

\paragraph{10.\;Nearest.}\mbox{}
\begin{lstlisting}[style=tmpl]
Q: Which node is spatially closest to {node.text}?
A: The node spatially closest to <|ref|>{node.text}<|/ref|><|box|>{node.bbox}<|/box|> is <|ref|>{nearest_node.text}<|/ref|><|box|>{nearest_node.bbox}<|/box|>.
\end{lstlisting}

\paragraph{11.\;Sorting.}

This task has two variants: top-to-bottom and left-to-right.

\begin{lstlisting}[style=tmpl]
Q: List all nodes in spatial order from top to bottom.
A: The top-to-bottom order of nodes is: <|ref|>{node_1.text}<|/ref|><|box|>{node_1.bbox}<|/box|> -> <|ref|>{node_2.text}<|/ref|><|box|>{node_2.bbox}<|/box|> -> ... -> <|ref|>{node_n.text}<|/ref|><|box|>{node_n.bbox}<|/box|>.
\end{lstlisting}
\begin{lstlisting}[style=tmpl]
Q: List all nodes in spatial order from left to right.
A: The left-to-right order of nodes is: <|ref|>{node_1.text}<|/ref|><|box|>{node_1.bbox}<|/box|> -> <|ref|>{node_2.text}<|/ref|><|box|>{node_2.bbox}<|/box|> -> ... -> <|ref|>{node_n.text}<|/ref|><|box|>{node_n.bbox}<|/box|>.
\end{lstlisting}

\paragraph{12.\;Coordinate.}

This task has two variants: grounding (text $\to$ box) and degrounding (box $\to$ text).

\begin{lstlisting}[style=tmpl]
Q: What are the coordinates of {node.text}?
A: The coordinates of {node.text} are <|ref|>{node.text}<|/ref|><|box|>{node.bbox}<|/box|>.
\end{lstlisting}
\begin{lstlisting}[style=tmpl]
Q: Which node is located in the region with coordinates {node.bbox}?
A: The node located in this region is <|ref|>{node.text}<|/ref|><|box|>{node.bbox}<|/box|>.
\end{lstlisting}

\end{document}